\documentclass[runningheads]{llncs}
\usepackage{graphicx}
\usepackage{float}  
\usepackage{subcaption} 

\usepackage{tikz}
\usepackage{comment}
\usepackage{amsmath,amssymb} 
\usepackage{color}
\usepackage[accsupp]{axessibility}  

\usepackage{multirow}
\usepackage{xcolor}
\usepackage{bbding}

\usepackage{amsmath}

\usepackage[pagebackref,breaklinks,colorlinks]{hyperref}

\begin{document}
\pagestyle{headings}
\mainmatter
\def\ECCVSubNumber{4669}  

\title{MOTR: End-to-End Multiple-Object Tracking with Transformer} 

\titlerunning{MOTR: End-to-End Multiple-Object Tracking with Transformer}
%
\author{Fangao Zeng\inst{1}\thanks{Equal contribution.} \and
  Bin Dong\inst{1}\textsuperscript{$\star$} \and
  Yuang Zhang\inst{2}\textsuperscript{$\star$} \and
  Tiancai Wang\inst{1}\thanks{Corresponding author. Email:  wangtiancai@megvii.com} \and \\
  Xiangyu Zhang\inst{1} \and
  Yichen Wei\inst{1}}
\authorrunning{F. Zeng et al.}
%
\institute{MEGVII Technology \and Shanghai Jiao Tong University}
\maketitle

\begin{abstract}
  Temporal modeling of objects is a key challenge in multiple-object tracking (MOT).
  Existing methods track by associating detections through motion-based and appearance-based similarity heuristics.
  The post-processing nature of association prevents end-to-end exploitation of temporal variations in video sequence.

  In this paper, we propose MOTR, which extends DETR \cite{carion2020detr} and introduces ``track query'' to model the tracked instances in the entire video. Track query is transferred and updated frame-by-frame to perform iterative prediction over time. We propose tracklet-aware label assignment to train track queries and newborn object queries.
  We further propose temporal aggregation network and collective average loss to enhance temporal relation modeling.
  Experimental results on DanceTrack show that MOTR significantly outperforms state-of-the-art method, ByteTrack~\cite{zhang2021bytetrack} by \textbf{6.5\%} on HOTA metric.
  On MOT17, MOTR outperforms our concurrent works, TrackFormer~\cite{Meinhardt2021trackformer} and TransTrack~\cite{transtrack}, on association performance.
  MOTR can serve as a stronger baseline for future research on temporal modeling and Transformer-based trackers.
  Code is available at \url{https://github.com/megvii-research/MOTR}.

  \keywords{Multiple-Object Tracking, Transformer, End-to-End}
\end{abstract}

\section{Introduction}
Multiple-object tracking (MOT) predicts the trajectories of instances in continuous image sequences \cite{wojke2017simple,bergmann2019twb}.
Most of existing methods separate the MOT temporal association into appearance and motion:
appearance variance is usually measured by pair-wise Re-ID similarity \cite{wang2019jde,zhang2020fairmot} while motion is modeled via IoU \cite{bochinski2017ioutracker} or Kalman Filtering \cite{bewley2016simple} heuristic.
These methods require similarity-based matching for post-processing, which becomes the bottleneck of temporal information flow across frames.
In this paper, we aim to introduce a fully end-to-end MOT framework featuring joint motion and appearance modeling.

\begin{figure}[t]
  \centering
  \includegraphics[width=\linewidth]{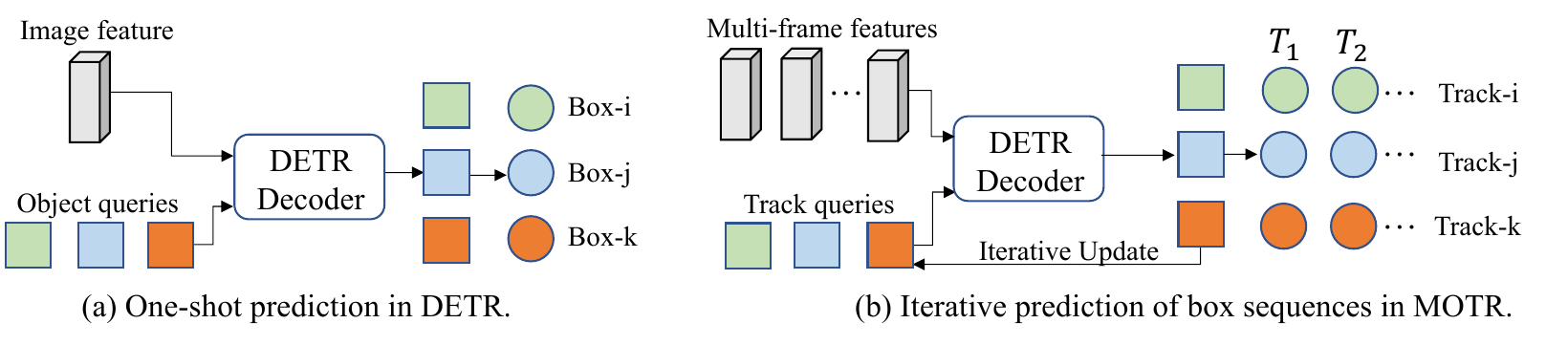}
  \caption{(a) DETR achieves end-to-end detection by interacting object queries with image features and performs one-to-one assignment between the updated queries and objects. (b) MOTR performs set of sequence prediction by updating the track queries. Each track query represents a track. Best viewed in color.}
  \label{method_compare_intro}
\end{figure}

Recently, DETR \cite{carion2020detr,zhu2020deformdetr} was proposed for end-to-end object detection. It formulates object detection as a set prediction problem. As shown in Fig.\;\ref{method_compare_intro}(a), object queries, served as a decoupled representation of objects, are fed into the Transformer decoder and interacted with the image feature to update their representation. Bipartite matching is further adopted to achieve one-to-one assignment between the object queries and ground-truths, eliminating post-processes, like NMS. Different from object detection, MOT can be regarded as a sequence prediction problem. The way to perform sequence prediction in the end-to-end DETR system is an open question.

Iterative prediction is popular in machine translation \cite{sutskever2014sequence,vaswani2017attention}. The output context is represented by a hidden state, and sentence features iteratively interact with the hidden state in the decoder to predict the translated words. Inspired by these advances in machine translation, we intuitively regard MOT as a problem of \emph{set of sequence prediction} since MOT requires a set of object sequences. Each sequence corresponds to an object trajectory. Technically, we extend object query in DETR to \emph{track query} for predicting object sequences. Track queries are served as the hidden states of object tracks. The representations of track queries are updated in the Transformer decoder and used to predict the object trajectory iteratively, as shown in Fig.\;\ref{method_compare_intro}(b). Specifically, track queries are updated through self-attention and cross-attention by frame features. The updated track queries are further used to predict the bounding boxes. The track of one object can be obtained from all predictions of one track query in different frames.

To achieve the goal above, we need to solve two problems: 1) track one object by one track query; 2) deal with newborn and terminated objects. To solve the first problem, we introduce tracklet-aware label assignment (TALA). It means that predictions of one track query are supervised by bounding box sequences with the same identity. To solve the second problem, we maintain a track query set of variable lengths. Queries of newborn objects are merged into this set while queries of terminated objects are removed. We name this process the entrance and exit mechanism.
In this way, MOTR does not require explicit track associations during inference. Moreover, the iterative update of track queries enables temporal modeling regarding both appearance and motion.

To enhance the temporal modeling, we further propose collective average loss (CAL) and temporal aggregation network (TAN). With the CAL, MOTR takes video clips as input during training. The parameters of MOTR are updated based on the overall loss calculated for the whole video clip. TAN introduces a shortcut for track query to aggregate the historical information from its previous states via the key-query mechanism in Transformer.

MOTR is a simple online tracker. It is easy to develop based on DETR with minor modifications on label assignment. It is a truly end-to-end MOT framework, requiring no post-processes, such as the track NMS or IoU matching employed in our concurrent works, TransTrack \cite{transtrack}, and TrackFormer \cite{Meinhardt2021trackformer}.
Experimental results on MOT17 and DanceTrack datasets show that MOTR achieves promising performance. On DanceTrack \cite{peize2021dance}, MOTR outperforms the state-of-the-art ByteTrack \cite{zhang2021bytetrack} by \textbf{6.5\%} on HOTA metric and \textbf{8.1\%} on AssA.

To summarize, our contributions are listed as below:
\begin{itemize}
  \item We present a fully end-to-end MOT framework, named MOTR. MOTR can implicitly learn the appearance and position variances in a joint manner.
  \item We formulate MOT as a problem of \emph{set of sequence prediction}. We generate track query from previous hidden states for iterative update and prediction.
  \item We propose tracklet-aware label assignment for one-to-one assignment between track queries and objects. An entrance and exit mechanism is introduced to deal with newborn and terminated tracks.
  \item We further propose CAL and TAN to enhance the temporal modeling.
\end{itemize}

\section{Related Work}

\noindent \textbf{Transformer-based Architectures.}
Transformer \cite{vaswani2017attention} was first introduced to aggregate information from the entire input sequence for machine translation. It mainly involves self-attention
and cross-attention mechanisms. Since that, it was gradually introduced to many fields, such as speech processing \cite{li2019neuralTransformer,chang2020speechTransformer} and computer vision \cite{wang2018nonlocal,camgoz2020signTransformer}. Recently, DETR \cite{carion2020detr} combined convolutional neural network (CNN), Transformer and bipartite matching to perform end-to-end object detection.  To achieve the fast convergence, Deformable DETR \cite{zhu2020deformdetr} introduced deformable attention module into Transformer encoder and Transformer decoder. ViT \cite{anonymous2021an} built a pure Transformer architecture for image classification.
Further, Swin Transformer \cite{liu2021swin} proposed shifted windowing scheme to perform self-attention within local windows, bringing greater efficiency. VisTR \cite{vistr2021} employed a direct end-to-end parallel sequence prediction framework to perform video instance segmentation.

\noindent \textbf{Multiple-Object Tracking.}
Dominant MOT methods mainly followed the tracking-by-detection paradigm \cite{bewley2016simple,leal2016learning,schulter2017deep,sharma2018beyond,wojke2017simple}. These approaches usually first employ object detectors to localize objects in each frame and then perform track association between adjacent frames to generate the tracking results. SORT \cite{bewley2016simple} conducted track association by combining Kalman Filter \cite{welch1995introduction} and Hungarian algorithm \cite{kuhn1955hungarian}. DeepSORT \cite{wojke2017simple} and Tracktor \cite{bergmann2019twb} introduced an extra cosine distance and compute the appearance similarity for track association. Track-RCNN \cite{shuai2020multi}, JDE \cite{wang2019jde} and FairMOT \cite{zhang2020fairmot} further added a Re-ID branch on top of object detector in a joint training framework, incorporating object detection and Re-ID feature learning. TransMOT \cite{chu2021transmot} builds a spatial-temporal graph transformer for association. Our concurrent works, TransTrack \cite{transtrack} and TrackFormer \cite{Meinhardt2021trackformer} also develop Transformer-based frameworks for MOT. For direct comparison with them, please refer to Sec.\;\ref{discussion}.

\noindent \textbf{Iterative Sequence Prediction.}
Predicting sequence via sequence-to-sequence (seq2seq) with encoder-decoder architecture is popular in machine translation \cite{sutskever2014sequence,vaswani2017attention} and text recognition \cite{shi2016end}. In seq2seq framework, the encoder network encodes the input into intermediate representation. Then, a hidden state with task-specific context information is introduced and iteratively interacted with the intermediate representation to generate the target sequence through the decoder network. The iterative decode process contains several iterations. In each iteration, hidden state decodes one element of target sequence.

\section{Method}
\subsection{Query in Object Detection}
DETR \cite{carion2020detr} introduced a fixed-length set of object queries to detect objects. Object queries are fed into the Transformer decoder and interacted with image features, extracted from Transformer encoder to update their representation. Bipartite matching is further adopted to achieve one-to-one assignment between the updated object queries and ground-truths. Here, we simply write the object query as ``detect query'' to specify the query used for object detection.

\subsection{Detect Query and Track Query}
\begin{figure}[t]
  \centering
  \includegraphics[width=\linewidth]{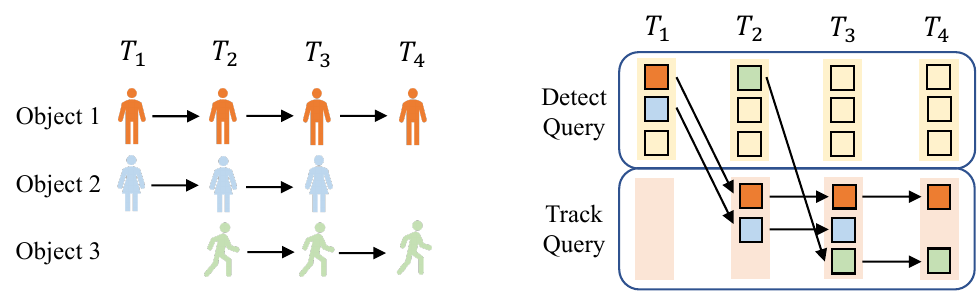}
  \caption{Update process of detect (object) queries and track queries under some typical MOT cases. Track query set is updated dynamically, and the length is variable. Track query set is initialized to be empty, and the detect queries are used to detect newborn objects. Hidden states of all detected objects are concatenated to produce track queries for the next frame. Track queries assigned to terminated objects are removed from the track query set.}
  \label{detect_track_query}
\end{figure}

When adapting DETR from object detection to MOT, two main problems arise: 1) how to track one object by one track query; 2) how to handle newborn and terminated objects.
We extend detect queries to track queries in this paper.
Track query set is updated dynamically, and the length is variable.
As shown in Fig.\;\ref{detect_track_query}, the track query set is initialized to be empty, and the detect queries in DETR are used to detect newborn objects (object 3 at $T_{2}$). Hidden states of detected objects produces track queries for the next frame; track queries assigned to terminated objects are removed from the track query set (object 2 at $T_{4}$).

\subsection{Tracklet-Aware Label Assignment}
\label{tala}
In DETR, one detect (object) query may be assigned to any object in the image since the label assignment is determined by performing bipartite matching between all detect queries and ground-truths. While in MOTR, detect queries are only used to detect newborn objects while track queries predict all tracked objects. Here, we introduce the tracklet-aware label assignment (TALA) to solve this problem.

Generally, TALA consists of two strategies. For detect queries, we modify the assignment strategy in DETR as \textbf{newborn-only}, where bipartite matching is conducted between the detect queries and the ground-truths of newborn objects. For track queries, we design an \textbf{target-consistent} assignment strategy. Track queries follow the same assignment of previous frames and are therefore excluded from the aforementioned bipartite matching.

Formally, we denote the predictions of track queries as $\widehat{Y}_{tr}$ and predictions of detect queries as $\widehat{Y}_{det}$.
$Y_{new}$ is the ground-truths of newborn objects.
The label assignment results for track queries and detect queries can be written as $\omega_{tr}$ and $\omega_{det}$.
For frame $i$, label assignment for detect queries is obtained from bipartite matching among detect queries and newborn objects, i.e.,
\begin{equation}
  \omega^{i}_{det}=\underset{\omega_{det}^{i} \in \Omega_{i}}{\mathrm{arg}\min}~\mathcal{L}(\widehat{Y}^{i}_{det}|_{\omega^{i}_{det}}, Y^{i}_{new}),
  \label{label_assign_new}
\end{equation}
where $\mathcal{L}$ is the pair-wise matching cost defined in DETR and $\Omega_{i}$ is the space of all bipartite matches among detect queries and newborn objects.
For track query assignment, we merge the assignments for newborn objects and tracked objects from the last frame, i.e., for $i>1$:
\begin{equation}
  \omega^{i}_{tr}= \omega^{i-1}_{tr} \cup \omega^{i-1}_{det}.
  \label{label_assign_tracked}
\end{equation}
For the first frame ($i=1$), track query assignment $\omega^{1}_{tr}$ is an empty set $\emptyset$ since there are no tracked objects for the first frame.
For successive frames ($i>1$), the track query assignment $\omega^{i}_{tr}$ is the concatenation of previous track query assignment $\omega^{i-1}_{tr}$ and newborn object assignment $\omega^{i-1}_{det}$.


In practice, the TALA strategy is simple and effective thanks to the powerful attention mechanism in Transformer.
For each frame, detect queries and track queries are concatenated and fed into the Transformer decoder to update their representation.
Detect queries will only detect newborn objects since query interaction by self-attention in the Transformer decoder will suppress detect queries that detect tracked objects.
This mechanism is similar to duplicate removal in DETR that duplicate boxes are suppressed with low scores.

\begin{figure}[t]
  \centering
  \includegraphics[width=\linewidth]{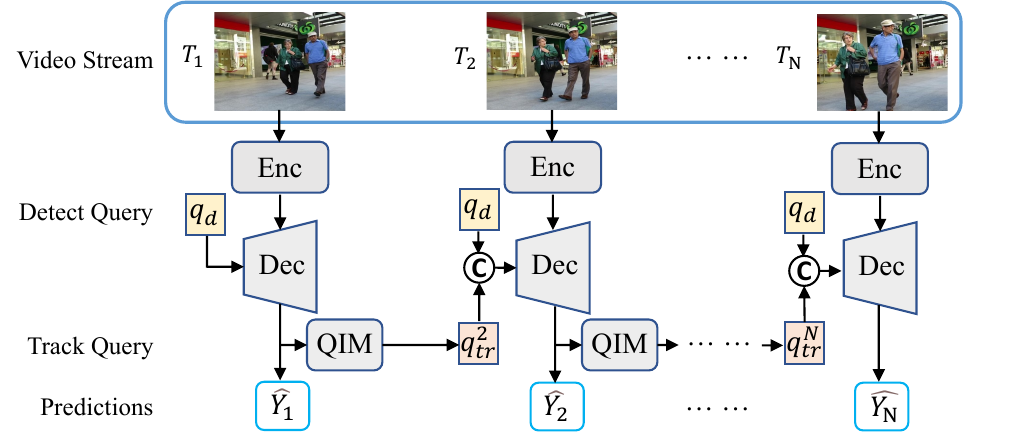}
  \caption{The overall architecture of MOTR. ``Enc'' represents a convolutional neural network backbone and the Transformer encoder that extracts image features for each frame. The concatenation of detect queries $q_{d}$ and track queries $q_{tr}$ is fed into the Deformable DETR decoder (Dec) to produce the hidden states. The hidden states are used to generate the prediction $\widehat{Y}$ of newborn and tracked objects. The query interaction module (QIM) takes the hidden states as input and produces track queries for the next frame.}
  \label{architecture}
\end{figure}

\subsection{MOTR Architecture}
\label{iterative_track}
The overall architecture of MOTR is shown in Fig.\;\ref{architecture}.
Video sequences are fed into the convolutional neural network (CNN) (e.g. ResNet-50 \cite{He2016Resnet}) and Deformable DETR \cite{zhu2020deformdetr} encoder to extract frame features.

For the first frame, there are no track query and we only feed the fixed-length learnable detect queries ($q_{d}$ in Fig.\;\ref{architecture}) into the Deformable DETR \cite{zhu2020deformdetr} decoder.
For successive frames, we feed the concatenation of track queries from the previous frame and the learnable detect queries into the decoder.
These queries interact with image feature in the decoder to generate the hidden state for bounding box prediction.
The hidden state is also fed into the query interaction module (QIM) to generate the track queries for the next frame.

During training phase, the label assignment for each frame is described in Sec. \ref{tala}.
All predictions of the video clip are collected into a prediction bank \{$\widehat{Y}_{1}$, $\widehat{Y}_{2}$, \dots, $\widehat{Y}_{N}$\}, and we use the proposed collective average loss (CAL) described in Sec.\;\ref{collective_loss} for supervision. During inference time, the video stream can be processed online and generate the prediction for each frame.

\subsection{Query Interaction Module}
\label{QIM}
In this section, we describe query interaction module (QIM). QIM includes object entrance and exit mechanism and temporal aggregation network (TAN).

\begin{figure}[t]
  \centering
  \includegraphics[width=\linewidth]{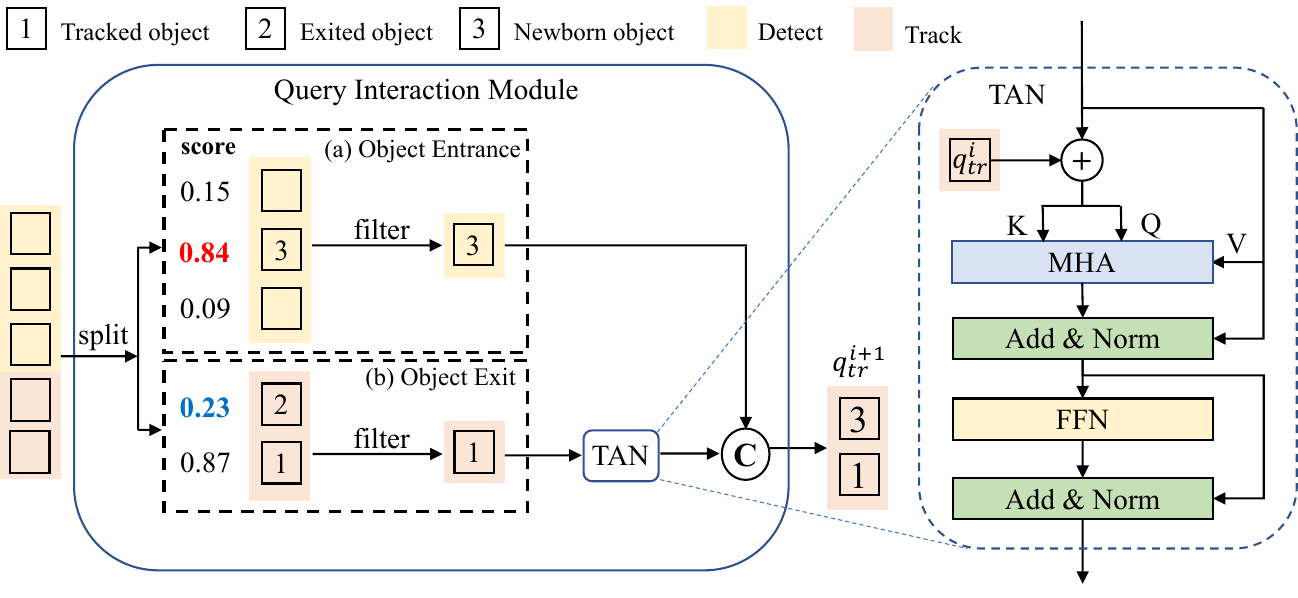}
  \caption{The structure of query interaction module (QIM). The inputs of QIM are the hidden state produced by Transformer decoder and the corresponding prediction scores. In the inference stage, we keep newborn objects and drop exited objects based on the confidence scores. Temporal aggregation network (TAN) enhances long-range temporal modeling.}
  \label{QIN}
\end{figure}

\noindent \textbf{Object Entrance and Exit.}
As mentioned above, some objects in video sequences may appear or disappear at intermediate frames. Here, we introduce the way we deal with the newborn and terminated objects in our method. For any frame, track queries are concatenated with the detect queries and input to the Transformer decoder, producing the hidden state (see the left side of Fig.\;\ref{QIN}).

During training, hidden states of terminated objects are removed if the matched objects disappeared in ground-truths or the intersection-over-union (IoU) between predicted bounding box and target is below a threshold of 0.5.
It means that the corresponding hidden states will be filtered if these objects disappear at current frame while the rest hidden states are reserved.
For newborn objects, the corresponding hidden states are kept based on the assignment of newborn object $\omega^{i}_{det}$ defined in Eq.\;\ref{label_assign_new}.

For inference, we use the predicted classification scores to determine appearance of newborn objects and disappearance of tracked objects, as shown in Fig.~\ref{QIN}.
For object queries, predictions whose classification scores are higher than the entrance threshold $\tau_{en}$ are kept while other hidden states are removed.
For track queries, predictions whose classification scores are lower than the exit threshold $\tau_{ex}$ for consecutive $M$ frames are removed while other hidden states are kept.

\noindent \textbf{Temporal Aggregation Network.}
Here, we introduce the temporal aggregation network (TAN) in QIM to enhance temporal relation modeling and provide contextual priors for tracked objects.

As shown in Fig.\;\ref{QIN}, the input of TAN is the filtered hidden state for tracked objects (object ``1'').
We also collect the track query $q_{tr}^{i}$ from the last frame for temporal aggregation.
TAN is a modified Transformer decoder layer. The track query from the last frame and the filtered hidden state are summed to be the key and query components of the multi-head self-attention (MHA).
The hidden state alone is the value component of MHA.
After MHA, we apply a feed-forward network (FFN) and the results are concatenated with the hidden state for newborn objects (object ``3'') to produce the track query set $q_{tr}^{i+1}$ for the next frame.

\subsection{Collective Average Loss}
\label{collective_loss}
Training samples are important for temporal modeling of track since MOTR learns temporal variances from data rather than hand-crafted heuristics like Kalman Filtering. Common training strategies, like training within two frames, fail to generate training samples of long-range object motion. Different from them, MOTR takes video clips as input. In this way, training samples of long-range object motion can be generated for temporal learning.

Instead of calculating the loss frame-by-frame, our collective average loss (CAL) collects the multiple predictions $\widehat{Y}=\{\widehat{Y}_{i}\}_{i=1}^{N}$. Then the loss within the whole video sequence is calculated by ground-truths $Y=\{Y_{i}\}_{i=1}^{N}$ and the matching results $\omega=\{\omega_{i}\}_{i=1}^{N}$.
CAL is the overall loss of the whole video sequence, normalized by the number of objects:
\begin{equation}
  \begin{split}
    \mathcal{L}_{o}(\widehat{Y}|_{\omega},Y) = \frac{\sum\limits_{n=1}^{N}(\mathcal{L}(\widehat{Y}^{i}_{tr}|_{\omega^{i}_{tr}},Y^{i}_{tr})+\mathcal{L}(\widehat{Y}^{i}_{det}|_{\omega^{i}_{det}},Y^{i}_{det})) }{\sum\limits_{n=1}^{N}(V_{i})}
  \end{split}
  \label{Eq8}
\end{equation}
where $V_{i}=V_{tr}^{i}+V_{det}^{i}$ denotes the total number of ground-truths objects at frame $i$. $V_{tr}^{i}$ and $V_{det}^{i}$ are the numbers of tracked objects and newborn objects at frame $i$, respectively. $\mathcal{L}$ is the loss of single frame, which is similar to the detection loss in DETR. The single-frame loss $\mathcal{L}$ can be formulated as:
\begin{equation}
  \begin{split}
    \mathcal{L}(\widehat{Y}_{i}|_{\omega_{i}},Y_{i}) = \lambda_{cls}\mathcal{L}_{cls} + \lambda_{l_{1}}\mathcal{L}_{l_{1}} + \lambda_{giou}\mathcal{L}_{giou}
  \end{split}
  \label{Eq9}
\end{equation}
where $\mathcal{L}_{cls}$ is the focal loss \cite{lin2017focalloss}. $\mathcal{L}_{l_{1}}$ denotes the L1 loss and $\mathcal{L}_{giou}$ is the generalized IoU loss \cite{rezatofighi2019giouloss}. $\lambda_{cls}$,  $\lambda_{l_{1}}$ and $\lambda_{giou}$ are the corresponding weight coefficients.

\subsection{Discussion}
\label{discussion}
Based on DETR, our concurrent works, TransTrack \cite{transtrack} and TrackFormer \cite{Meinhardt2021trackformer} also develop the Transformer-based frameworks for MOT. However, our method shows large differences compared to them:

\textbf{TransTrack} models a full track as a combination of several independent short tracklets.
Similar to the track-by-detection paradigm, TransTrack decouples MOT as two sub-tasks:
1) detect object pairs as short tracklets within two adjacent frames;
2) associate short tracklets as full tracks by IoU-matching.
While for MOTR, we model a full track in an end-to-end manner through the iterative update of track query, requiring no IoU-matching.

\textbf{TrackFormer} shares the idea of track query with us.
However, TrackFormer still learns within two adjacent frames.
As discussed in Sec.\;\ref{collective_loss}, learning within short-range will result in relatively weak temporal learning.
Therefore, TrackFormer employs heuristics, such as Track NMS and Re-ID features, to filter out duplicate tracks.
Different from TrackFormer, MOTR learns stronger temporal motion with CAL and TAN, removing the need of those heuristics.
For direct comparison with TransTrack and TrackFormer, please refer to the Table\;\ref{tab:properties}.

Here, we clarify that we started this work independently long before TrackFormer and TransTrack appear on arXiv.
Adding that they are not formally published, we treat them as \emph{concurrent and independent} works instead of \emph{previous} works on which our work is built upon.

\begin{table}[t]
  \centering
  \begin{minipage}[c]{0.48\textwidth}
    \centering
    \captionof{table}{Comparison with other MOT methods based on Transformer.}
    \resizebox{1.0\linewidth}{!}{
      \begin{tabular}{l|cccc}
        \hline
        Method                                      & IoU match  & NMS        & ReID       \\
        \hline
        TransTrack \cite{transtrack}                & \Checkmark &            &            \\
        TrackFormer \cite{Meinhardt2021trackformer} &            & \Checkmark & \Checkmark \\
        MOTR (ours)                                 &            &            &            \\
        \hline
      \end{tabular}
    }
    \label{tab:properties}
  \end{minipage}
  \hspace{0.02\textwidth}
  \begin{minipage}[c]{0.48\textwidth}
    \centering
    \captionof{table}{Statistics of chosen datasets for evaluation.}
    \resizebox{1.0\linewidth}{!}{
      \begin{tabular}{l|cccc}
        \hline
        Datasets                         & Class & Frame & Video & ID   \\
        \hline
        DanceTrack \cite{peize2021dance} & 1     & 106k  & 100   & 990  \\
        MOT17 \cite{milan2016mot16}      & 1     & 11k   & 14    & 1342 \\
        BDD100K \cite{bdd100k}           & 8     & 318k  & 1400  & 131k \\
        \hline
      \end{tabular}
    }
    \label{tab:datasets}
  \end{minipage}
\end{table}

\section{Experiments}

\subsection{Datasets and Metrics}
\label{datasetsandmetrics}

\noindent \textbf{Datasets.}
For comprehensive evaluation, we conducted experiments on three datasets: DanceTrack \cite{peize2021dance}, MOT17 \cite{milan2016mot16}, and BDD100k \cite{bdd100k}.
MOT17 \cite{milan2016mot16} contains 7 training sequences and 7 test sequences.
DanceTrack \cite{peize2021dance} is a recent multi-object tracking dataset featuring uniform appearance and diverse motion. It contains more videos for training and evaluation thus providing a better choice to verify the tracking performance.
BDD100k \cite{bdd100k} is an autonomous driving dataset with an MOT track featuring multiple object classes.
For more details, please refer to the statistics of datasets, shown in Table \ref{tab:datasets}.

\noindent \textbf{Evaluation Metrics.}
We follow the standard evaluation protocols to evaluate our method.
The common metrics include Higher Order Metric for Evaluating Multi-object Tracking \cite{hota2021} (HOTA, AssA, DetA), Multiple-Object Tracking Accuracy (MOTA), Identity Switches (IDS) and Identity F1 Score (IDF1).

\subsection{Implementation Details}
Following the settings in CenterTrack \cite{zhou2020centertrack}, MOTR adopts several data augmentation methods, such as random flip and random crop. The shorter side of the input image is resized to 800 and the maximum size is restricted to 1536. The inference speed on Tesla V100 at this resolution is about 7.5 FPS. We sample keyframes with random intervals to solve the problem of variable frame rates. Besides, we erase the tracked queries with the probability $p_{drop}$ to generate more samples for newborn objects and insert track queries of false positives with the probability $p_{insert}$ to simulate the terminated objects.
All the experiments are conducted on PyTorch with 8 NVIDIA Tesla V100 GPUs. We also provide a memory-optimized version that can be trained on NVIDIA 2080 Ti GPUs.

We built MOTR upon Deformable-DETR \cite{zhu2020deformdetr} with ResNet50 \cite{He2016Resnet} for fast convergence. The batch size is set to 1 and each batch contains a video clip of 5 frames.
We train our model with the AdamW optimizer with the initial learning rate of $2.0\cdot 10^{-4}$. For all datasets, we initialize MOTR with the official Deformable DETR \cite{zhu2020deformdetr} weights pre-trained on the COCO \cite{coco14} dataset.
On \textbf{MOT17}, we train MOTR for 200 epochs and the learning rate decays by a factor of 10 at the $100^\text{th}$ epoch.
For state-of-the-art comparison, we train on the joint dataset (MOT17 training set and CrowdHuman \cite{shao2018crowdhuman} val set). For $\sim$5k static images in CrowdHuman val set, we apply random shift as in \cite{zhou2020centertrack} to generate video clips with pseudo tracks. The initial length of video clip is 2 and we gradually increase it to 3,4,5 at the $50^\text{th}$,$90^\text{th}$,$150^\text{th}$ epochs, respectively. The progressive increment of video clip length improves the training efficiency and stability. For the ablation study, we train MOTR on the MOT17 training set without using the CrowdHuman dataset and validate on the 2DMOT15 training set.
On \textbf{DanceTrack}, we train for 20 epochs on the train set and learning rate decays at the $10^\text{th}$ epoch. We gradually increase the clip length from 2 to 3,4,5 at the $5^\text{th}$,$9^\text{th}$,$15^\text{th}$ epochs.
On \textbf{BDD100k}, we train for 20 epochs on the train set and learning rate decays at the $16^\text{th}$ epoch. We gradually increase the clip length from 2 to 3 and 4 at the $6^\text{th}$ and $12^\text{th}$ epochs.

\subsection{State-of-the-art Comparison on MOT17}
Table\;\ref{tab_compare_sota} compares our approach with state-of-the-art methods on MOT17 test set. We mainly compare MOTR with our concurrent works based on Transformer: TrackFormer \cite{Meinhardt2021trackformer} and TransTrack \cite{transtrack}. Our method gets higher IDF1 scores, surpassing TransTrack and TrackFormer by 4.5\%. The performance of MOTR on the HOTA metric is much higher than TransTrack by 3.1\%. For the MOTA metric, our method achieves much better performance than TrackFormer (71.9\% \textit{vs.} 65.0\%). Interestingly, we find that the performance of TransTrack is better than our MOTR on MOTA. We suppose the decoupling of detection and tracking branches in TransTrack indeed improves the object detection performance. While in MOTR, detect and track queries are learned through a shared Transformer decoder. Detect queries are suppressed on detecting tracked objects, limiting the detection performance on newborn objects.

If we compare the performance with other state-of-the-art methods, like ByteTrack~\cite{zhang2021bytetrack}, it shows that MOTR is \textbf{frustratingly inferior} to them on the MOT17 dataset. Usually, state-of-the-art performance on the MOT17 dataset is dominated by trackers with good detection performance to cope with various appearance distributions. Also, different trackers tend to employ different detectors for object detection. It is pretty difficult for us to fairly verify the motion performance of various trackers. Therefore, we argue that the MOT17 dataset alone is \textbf{not enough} to fully evaluate the tracking performance of MOTR.
We further evaluate the tracking performance on DanceTrack~\cite{peize2021dance} dataset with uniform appearance and diverse motion, as described next.

\begin{table}[t]
  \centering
  \caption{Performance comparison between MOTR and existing methods on the MOT17 dataset under the private detection protocols. The number is marked in bold if it is the best among the Transformer-based methods.}
  \setlength{\tabcolsep}{2.5mm}{
    \begin{tabular}{l|cccccc}
      \hline
      Methods                                     & HOTA$\uparrow$ & AssA$\uparrow$ & DetA$\uparrow$ & IDF1$\uparrow$ & MOTA$\uparrow$ & IDS$\downarrow$ \\
      \hline
      \textit{CNN-based:}                         &                &                &                &                &                &                 \\
      Tracktor++\cite{bergmann2019twb}            & 44.8           & 45.1           & 44.9           & 52.3           & 53.5           & 2072            \\
      CenterTrack\cite{zhou2020centertrack}       & 52.2           & 51.0           & 53.8           & 64.7           & 67.8           & 3039            \\
      TraDeS \cite{trades2021}                    & 52.7           & 50.8           & 55.2           & 63.9           & 69.1           & 3555            \\
      QDTrack \cite{quasisense2021}            & 53.9           & 52.7           & 55.6           & 66.3           & 68.7           & 3378            \\
      GSDT \cite{wang2021joint}                   & 55.5           & 54.8           & 56.4           & 68.7           & 66.2           & 3318            \\
      FairMOT\cite{zhang2020fairmot}              & 59.3           & 58.0           & 60.9           & 72.3           & 73.7           & 3303            \\
      CorrTracker \cite{wang2021multiple}         & 60.7           & 58.9           & 62.9           & 73.6           & 76.5           & 3369            \\
      GRTU \cite{wang2021general}                 & 62.0           & 62.1           & 62.1           & 75.0           & 74.9           & 1812            \\
      MAATrack \cite{stadler2022modelling}        & 62.0           & 60.2           & 64.2           & 75.9           & 79.4           & 1452            \\
      ByteTrack \cite{zhang2021bytetrack}         & 63.1           & 62.0           & 64.5           & 77.3           & 80.3           & 2196            \\
      \hline
      \textit{Transformer-based:}                 &                &                &                &                &                &                 \\
      TrackFormer \cite{Meinhardt2021trackformer} & /              & /              & /              & 63.9           & 65.0           & 3528            \\
      TransTrack\cite{transtrack}                 & 54.1           & 47.9           & \textbf{61.6}  & 63.9           & \textbf{74.5}  & 3663            \\
      MOTR (ours)                                 & \textbf{57.8}  & \textbf{55.7}  & 60.3           & \textbf{68.6}  & 73.4           & \textbf{2439}   \\
      \hline
    \end{tabular}}
  \label{tab_compare_sota}
\end{table}

\subsection{State-of-the-art Comparison on DanceTrack}
Recently, DanceTrack~\cite{peize2021dance}, a dataset with uniform appearance and diverse motion, is introduced (see Tab.~\ref{tab:datasets}). It contains much more videos for evaluation and provides a better choice to verify the tracking performance.
We further conduct the experiments on the DanceTrack dataset and perform the performance comparison with state-of-the-art methods in Tab.~\ref{tab_compare_dancetrack}. It shows that MOTR achieves much better performance on DanceTrack dataset.
Our method gets a much higher HOTA score, surpassing ByteTrack by 6.5\%. For the AssA metric, our method also achieves much better performance than ByteTrack (40.2\% \textit{vs.} 32.1\%). While for the DetA metric, MOTR is inferior to some state-of-the-art methods. It means that MOTR performs well on temporal motion learning while the detection performance is not that good. The large improvements on HOTA are mainly from the temporal aggregation network and collective average loss.

\begin{table}[t]
  \centering
  \caption{Performance comparison between MOTR and existing methods on the DanceTrack\cite{peize2021dance} dataset. Results for existing methods are from DanceTrack \cite{peize2021dance}.}
  \setlength{\tabcolsep}{2.5mm}{
    \begin{tabular}{l|ccccc}
      \hline
      Methods                                & HOTA          & AssA          & DetA          & MOTA          & IDF1          \\\hline
      CenterTrack \cite{zhou2020centertrack} & 41.8          & 22.6          & \textbf{78.1} & 86.8          & 35.7          \\
      FairMOT \cite{zhang2020fairmot}        & 39.7          & 23.8          & 66.7          & 82.2          & 40.8          \\
      QDTrack \cite{quasisense2021}          & 45.7          & 29.2          & 72.1          & 83.0          & 44.8          \\
      TransTrack \cite{transtrack}           & 45.5          & 27.5          & 75.9          & 88.4          & 45.2          \\
      TraDes \cite{trades2021}               & 43.3          & 25.4          & 74.5          & 86.2          & 41.2          \\
      ByteTrack \cite{zhang2021bytetrack}    & 47.7          & 32.1          & 71.0          & \textbf{89.6} & \textbf{53.9} \\\hline
      MOTR (ours)                            & \textbf{54.2} & \textbf{40.2} & 73.5          & 79.7          & 51.5          \\\hline
    \end{tabular}
  }
  \label{tab_compare_dancetrack}
\end{table}

\subsection{Generalization on Multi-Class Scene}
Re-ID based methods, like FairMOT~\cite{zhang2020fairmot}, tend to regard each tracked object (e.g., person) as a class and associate the detection results by the feature similarity. However, the association will be difficult when the number of tracked objects is very large. Different from them, each object is denoted as one track query in MOTR and the track query set is of dynamic length. MOTR can easily deal with the multi-class prediction problem, by simply modifying the class number of the classification branch. To verify the performance of MOTR on multi-class scenes, we further conduct the experiments on the BDD100k dataset (see Tab.~\ref{tab:bdd100k}). Results on bdd100k validation set show that MOTR performs well on multi-class scenes and achieves promising performance with fewer ID switches.

\begin{table}[t]
  \centering
  \caption{Performance comparison between MOTR and existing methods on the BDD100k\cite{bdd100k} validation set.}
  \setlength{\tabcolsep}{2.5mm}{
    \begin{tabular}{l|ccc}
      \hline
      Methods                    & mMOTA         & mIDF1         & IDSw          \\
      \hline
      Yu \emph{et al.} \cite{bdd100k} & 25.9          & \textbf{44.5} & 8315         \\
      DeepBlueAI \cite{bdd100k2020}  & 26.9         & $/$          & 13366         \\
      MOTR (ours)                & \textbf{32.0} & 43.5          & \textbf{3493} \\
      \hline
    \end{tabular}
  }
  \label{tab:bdd100k}
\end{table}

\subsection{Ablation Study}

\begin{table}[t]
  \centering
  \caption{Ablation studies on our proposed MOTR. All experiments use the single-level C5 feature in ResNet50.}
  \subfloat[The effect of our contributions. TrackQ: track query. TAN: temporal aggregation network. CAL: collective average loss. \label{abs-a}]{
    \resizebox{0.51\linewidth}{!}{
      \setlength{\tabcolsep}{1mm}{
        \begin{tabular}{cccccc}
          \hline
          TrackQ     & TAN        & CAL        & MOTA$\uparrow$ & IDF1$\uparrow$ & IDS$\downarrow$ \\
          \hline
          ~          & ~          & ~          & -              & 1.2            & 33198           \\
          \checkmark & ~          & ~          & 37.1           & 49.8           & 562             \\
          \checkmark & \checkmark & ~          & 44.9           & 63.4           & 257             \\
          \checkmark & ~          & \checkmark & 47.5           & 56.1           & 417             \\
          \checkmark & \checkmark & \checkmark & \textbf{53.2}  & \textbf{70.5}  & \textbf{155}    \\
          \hline
        \end{tabular}
      }}}
  \hspace{0.02\linewidth}
  \subfloat[The impact of increasing video clip length in Collective Average Loss during training on tracking performance. \label{abs-b}]{
    \resizebox{0.44\linewidth}{!}{
      \setlength{\tabcolsep}{2.2mm}{
        \begin{tabular}{cccccc}
          \hline
          Length     & MOTA$\uparrow$ & IDF1$\uparrow$ & IDS$\downarrow$ \\
          \hline
          2          & 44.9           & 63.4           & 257             \\
          3          & 51.6           & 59.4           & 424             \\
          4          & 50.6           & 64.0           & 314             \\
          \textbf{5} & \textbf{53.2}  & \textbf{70.5}  & \textbf{155}    \\
          \hline
        \end{tabular}
      }}}\vspace{2mm}\\
  \subfloat[Analysis on random track query erasing probability $p_{drop}$ during training. \label{abs-c}]{
    \resizebox{0.47\linewidth}{!}{
      \setlength{\tabcolsep}{3mm}{
        \begin{tabular}{cccccc}
          \hline
          $p_{drop}$   & MOTA$\uparrow$ & IDF1$\uparrow$ & IDS$\downarrow$ \\
          \hline
          5e-2         & 49.0           & 60.4           & 411             \\
          \textbf{0.1} & \textbf{53.2}  & \textbf{70.5}  & \textbf{155}    \\
          0.3          & 51.1           & 69.0           & 180             \\
          0.5          & 48.5           & 62.0           & 302             \\
          \hline
        \end{tabular}
      }}}
  \hspace{0.02\linewidth}\vspace{-1mm}
  \subfloat[Effect of random false positive inserting probability $p_{insert}$ during training. \label{abs-d}]{
    \resizebox{0.49\linewidth}{!}{
      \setlength{\tabcolsep}{3mm}{
        \begin{tabular}{cccccc}
          \hline
          $p_{insert}$ & MOTA$\uparrow$ & IDF1$\uparrow$ & IDS$\downarrow$ \\
          \hline
          0.1          & 51.2           & \textbf{71.7}  & \textbf{148}    \\
          \textbf{0.3} & \textbf{53.2}  & 70.5           & 155             \\
          0.5          & 52.1           & 62.0           & 345             \\
          0.7          & 50.7           & 57.7           & 444             \\
          \hline
        \end{tabular}
      }}}\vspace{2mm}\\
  \subfloat[The exploration of different combinations of $\tau_{ex}$ and $\tau_{en}$ in QIM network.
    \label{abs-e}]{
    \resizebox{0.54\linewidth}{!}{
      \setlength{\tabcolsep}{1mm}{
        \begin{tabular}{l|ccc|ccc}
          \hline
          $\tau_{ex}$     & 0.6  & 0.6           & 0.6          & 0.5           & 0.6           & 0.7  \\
          $\tau_{en}$     & 0.7  & 0.8           & 0.9          & 0.8           & 0.8           & 0.8  \\
          \hline
          MOTA$\uparrow$  & 52.7 & \textbf{53.2} & 53.1         & \textbf{53.5} & 53.2          & 52.8 \\
          IDF1$\uparrow$  & 69.8 & \textbf{70.5} & 70.1         & \textbf{70.5} & 70.5 & 68.3 \\
          IDS$\downarrow$ & 181  & 155           & \textbf{142} & \textbf{153}  & 155           & 181  \\
          \hline
        \end{tabular}
      }}}
  \hspace{0.02\linewidth}\vspace{-1mm}
  \subfloat[The effect of random sampling interval on tracking performance.
    \label{abs-f}]{
    \resizebox{0.42\linewidth}{!}{
      \setlength{\tabcolsep}{1mm}{
        \begin{tabular}{ccccc}
          \hline
          Intervals   & MOTA$\uparrow$ & IDF1$\uparrow$ & IDS$\downarrow$ \\
          \hline
          3           & 53.2           & 64.8           & 218             \\
          5           & 50.8           & 62.8           & 324             \\
          \textbf{10} & \textbf{53.2}  & \textbf{70.5}  & \textbf{155}    \\
          12          & 53.1           & 69             & 158             \\
          \hline
        \end{tabular}
      }}} \vspace{-1mm}
  \label{tab:abs}
\end{table}


\noindent \textbf{MOTR Components.}
Table\;\ref{abs-a} shows the impact of integrating different components. Integrating our components into the baseline can gradually improve overall performance. Using only object query of as original leads to numerous IDs since most objects are treated as entrance objects. With track query introduced, the baseline is able to handle tracking association and improve IDF1 from 1.2 to 49.8. Further, adding TAN to the baseline improves MOTA by 7.8\% and IDF1 by 13.6\%. When using CAL during training, there are extra 8.3\% and 7.1\% improvements in MOTA and IDF1, respectively. It demonstrates that TAN combined with CAL can enhance the learning of temporal motion.

\begin{figure}[t]
  \centering
  \includegraphics[width=\textwidth]{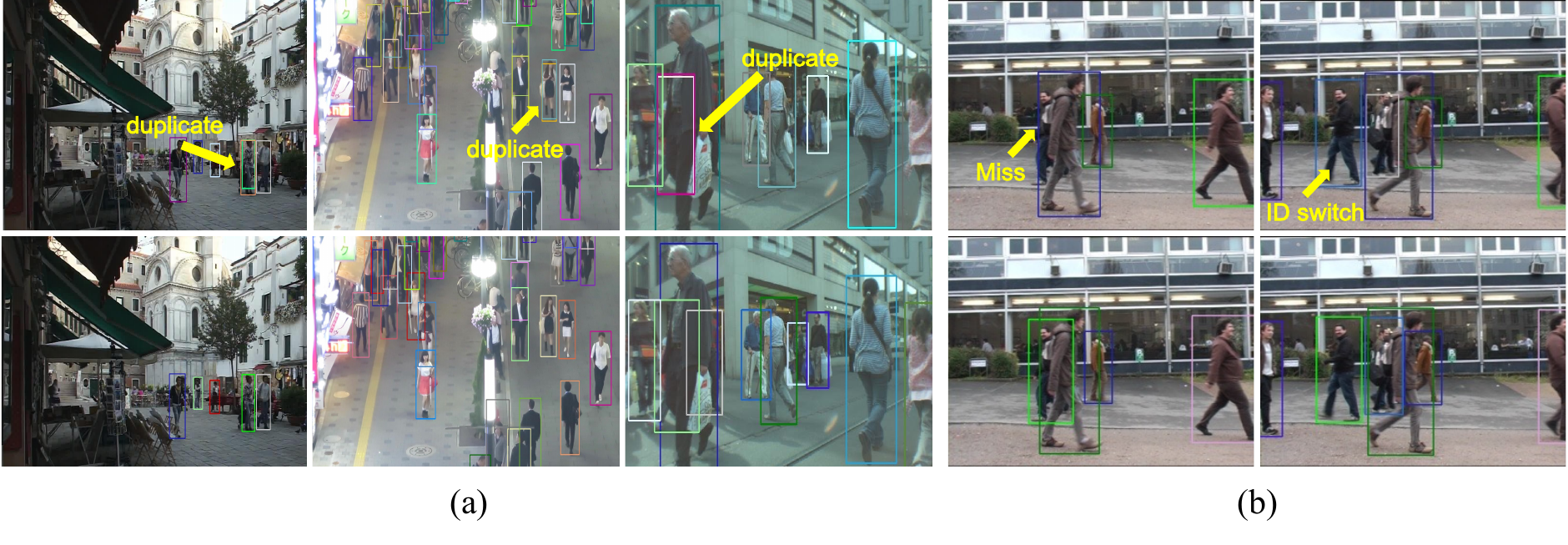}
  \vspace{-0.8cm}
  \caption{The effect of CAL on solving (a) duplicated boxes and (b) ID switch problems. Top and bottom rows are the tracking results without and with CAL, respectively.}
  \label{vis_duplicate}
\end{figure}

\noindent \textbf{Collective Average Loss.} Here, we explored the impact of video sequence length on the tracking performance in CAL. As shown in Table\;\ref{abs-b}, when the length of the video clip gradually increases from 2 to 5, MOTA and IDF1 metrics are improved by 8.3\% and 7.1\%, respectively. Thus, multi-frame CAL can greatly boost the tracking performance. We explained that multiple frames CAL can help the network to handle some hard cases such as occlusion scenes. We observed that duplicated boxes, ID switches, and object missing in occluded scenes are significantly reduced. To verify it, we provide some visualizations in Fig.\;\ref{vis_duplicate}.

\noindent \textbf{Erasing and Inserting Track Query.} In MOT datasets, there are few training samples for two cases: entrance objects and exit objects in video sequences. Therefore, we adopt track query erasing and inserting to simulate these two cases with probability $p_{drop}$ and $p_{insert}$, respectively. Table\;\ref{abs-c} reports the performance using different value of $p_{drop}$ during training. MOTR achieves the best performance when $p_{drop}$ is set to 0.1. Similar to the entrance objects, track queries transferred from the previous frame, whose predictions are false positives, are inserted into the current frame to simulate the case of object exit. In Table\;\ref{abs-d}, we explore the impact on tracking performance of different $p_{insert}$. When progressively increasing $p_{insert}$ from 0.1 to 0.7, our MOTR achieves the highest score on MOTA when $p_{insert}$ is set to 0.3
while the IDF1 score is decreasing.

\noindent \textbf{Object Entrance and Exit Threshold.}
Table\;\ref{abs-e} investigates the impact of different combination of object entrance threshold $\tau_{en}$ and exit threshold $\tau_{ex}$ in QIM. As we vary the object entrance threshold $\tau_{en}$, we can see that the performance is not that sensitive to $\tau_{en}$ (within 0.5\% on MOTA) and using an entrance threshold of 0.8 produces relatively better performance. We also further conduct experiments by varying the object exit threshold $\tau_{ex}$. It is shown that using a threshold of 0.5 results in slightly better performance than that of 0.6. In our practice, $\tau_{en}$ with 0.6 shows better performance on the MOT17 test set.

\noindent \textbf{Sampling Interval.}
In Table\;\ref{abs-f}, we evaluate the effect of random sampling interval on tracking performance during training. When the sampling interval increases from 2 to 10, the IDS decreases significantly from 209 to 155. During training, the network is easy to fall into a local optimal solution when the frames are sampled in a small interval. Appropriate increment on sampling interval can simulate real scenes. When the random sampling interval is greater than 10, the tracking framework fails to capture such long-range dynamics, leading to relatively worse tracking performance.

\section{Limitations}
MOTR, an online tracker, achieves end-to-end multiple-object tracking. It implicitly learns the appearance and position variances in a joint manner thanks to the DETR architecture as well as the tracklet-aware label assignment. However, it also has several shortcomings. First, the performance of detecting newborn objects is far from satisfactory (the result on the MOTA metric is not good enough). As we analyzed above, detect queries are suppressed on detecting tracked objects, which may go against the nature of object query and limits the detection performance on newborn objects. Second, the query passing in MOTR is performed frame-by-frame, limiting the efficiency of model learning during training. In our practice, the parallel decoding in VisTR \cite{vistr2021} fails to deal with the complex scenarios in MOT. Solving these two problems above will be an important research topic for Transformer-based MOT frameworks.

\subsubsection{Acknowledgements:} This research was supported by National Key R\&D Program of China (No. 2017YFA0700800) and Beijing Academy of Artificial Intelligence (BAAI).

%
%
\bibliographystyle{splncs04}
\bibliography{egbib}
\end{document}